%% file: main.tex
\documentclass{texsupport.IOS-Book-Article-master/IOS-Book-Article}

\usepackage{graphicx}
\usepackage[noabbrev, nameinlink]{cleveref}
\usepackage{tcolorbox}
\usepackage{listings}
\usepackage{mathptmx}
\usepackage{soul}\setuldepth{article}

\def\hb{\hbox to 11.5 cm{}}

\definecolor{blue-frame}{RGB}{108,142,191}
\definecolor{blue-background}{RGB}{218,232,252}

\definecolor{codegreen}{rgb}{0,0.6,0}
\definecolor{codegray}{rgb}{0.5,0.5,0.5}
\definecolor{codepurple}{rgb}{0.58,0,0.82}
\definecolor{backcolour}{rgb}{0.95,0.95,0.92}

\lstdefinestyle{codestyle}{
    backgroundcolor=\color{blue-background},
    commentstyle=\color{codepurple},
    frame=leftline,
    breakatwhitespace=false,         
    breaklines=true,                 
    captionpos=t,                    
    keepspaces=true,
    showstringspaces=false,
    numbersep=5pt,             
    tabsize=2,
    basicstyle=\ttfamily\scriptsize,
    rulecolor=\color{blue-frame}
}
\begin{document}
    \pagestyle{headings}
    \def\thepage{}
    \begin{frontmatter}
        \title{Towards Machine-Generated Code for the Resolution of User Intentions}
        
        \author[A]{\fnms{Justus} \snm{Flerlage}\orcid{0009-0007-2929-3408}
        \thanks{Corresponding Author: Justus Flerlage, j.flerlage@tu-berlin.de}},
        \author[A]{\fnms{Ilja} \snm{Behnke}}\orcid{0000-0002-2437-8994} and
        \author[A]{\fnms{Odej} \snm{Kao}}\orcid{0000-0001-6454-6799}
        \address[A]{Distributed and Operating Systems Group\\Technische Universität Berlin\\Berlin, Germany}
    
        \begin{abstract}
            \input{section/abstract}
        \end{abstract}
        
    \end{frontmatter}
    
    \input{section/introduction}
    \input{section/related-work}
    \input{section/preliminary-considerations}
    \input{section/method}
    \input{section/results}
    \input{section/conclusion}
    
    \bibliographystyle{texsupport.IOS-Book-Article-master/vancouver}
    \bibliography{references}
\end{document}

%% file: section/abstract.tex
The growing capabilities of Artificial Intelligence (AI), particularly Large Language Models (LLMs), prompt a reassessment of the interaction mechanisms between users and their devices. Currently, users are required to use a set of high-level applications to achieve their desired results. However, the advent of AI may signal a shift in this regard, as its capabilities have generated novel prospects for user-provided intent resolution through the deployment of model-generated code. This development represents a significant progression in the realm of hybrid workflows, where human and artificial intelligence collaborate to address user intentions, with the former responsible for defining these intentions and the latter for implementing the solutions to address them. In this paper, we investigate the feasibility of generating and executing workflows through code generation that results from prompting an LLM with a concrete user intention, and a simplified application programming interface for a GUI-less operating system. We provide an in-depth analysis and comparison of various user intentions, the resulting code, and its execution. The findings demonstrate the general feasibility of our approach and that the employed LLM, GPT-4o-mini, exhibits remarkable proficiency in the generation of code-oriented workflows in accordance with provided user intentions.

%% file: section/introduction.tex
\section{Introduction}
The current user interfaces on smartphones and other devices implement an action-oriented paradigm, where the user transforms their own intention into a workflow of commands and executes those by clicking on the corresponding entities, mostly apps, web services or other activity representations. For example, the request "reschedule my appointment for tonight" appearing to be a relatively straightforward task, actually necessitates a plethora of sophisticated activities. The initial action for the user is to launch the calendar application in order to ascertain which appointment with whom requires rescheduling. Subsequently, the user is required to access the contacts application to ascertain the telephone number of the appointment participant, open a channel of communication and commence negotiations regarding an alternative appointment. This simple example illustrates the burdensome obligation for the user to ensure a consistency between the individual steps, gather, adapt, and transfer the data between the workflow stages, find ad-hoc alternatives in case of failures, and repeat the procedure until a satisfactory result is achieved. Thus, the intelligence for the workflow design as well as the orchestration of the execution are located at the user. Moreover, the efficiency of the process and the user experience heavily depend on the user experience and skills using electronic devices.

With the advent of LLMs, an opportunity arises to gain assistance with the design and execution of such everyday workflows. Currently available LLMs are able to understand natural language, identify the intention in the submitted phrase, and provide answers or perform actions such as translation or summary of given documents \cite{jin2024comprehensive}. However, LLMs offer a further opportunity to decompose the intent into actionable steps and thus design a workflow in a similar fashion as human users. This capacity facilitates a transition from an action-based user interface to an intention-based user interface. In this new paradigm, an LLM assumes the role of an input processor for intents articulated in natural language, a semantic analyst and an intention interpreter, a workflow designer, and an execution orchestrator. This transition reduces the effort required from the device user, ultimately resulting in a hybrid interaction and collaboration between the user and the artificial intelligence, where the user articulates intentions and the artificial intelligence actualises these.

The key to the effective utilisation of such LLMs is to embed them into the local device. Currently, due to the substantial resource requirements of the inference step, it is necessary to perform this process on remote infrastructure. Target devices for this type of intent resolution can be future smartphones in combination with wearables such as the currently introduced camera-equipped glasses. Removing large smartphone screens creates significant challenges, but also allows a major reduction in size and leads to lower energy consumption by 80 \% \cite{Carroll2017UnderstandingAR}. Users may formulate their intentions in natural language and receive acknowledgments and a resulting summary once the task is processed. A detailed and urgent view on the incoming information can be projected on nearby screens or on the glasses. Replacing the screen with camera-glasses will allow a two-way visual communication: camera input along with the position data, surrounding  noise, and other sensor information will provide valuable multimodal information to be processed by the LLM and played back enriched with augmented information. 

In this paper we present the first step on the path to such a GUI-less operating system. We investigate, whether existing LLMs can be used for the recognition and transformation of intentions into workflows of application activities to fulfill them. We employ a LLM to process typical examples of intentions, subsequently extracting code with sequences of operations as the output, which functions as an abstraction of finite state machines. The employed LLM for this is OpenAI's \emph{GPT-4o-mini} \cite{achiam2023gpt} model accessed via a paid plan for programmatic access. We present a proof of concept that demonstrates the feasibility of using LLMs to generate and execute code that addresses a simplified application programming interface (API) for resolving user intentions. Additionally, we evaluate the quality and performance of the generated results.

In \Cref{section:related-work} of this paper, the related work in the domain of LLMs is presented. This is followed by the introduction of preliminary considerations in \Cref{section:preliminary-considerations}. \Cref{section:method} introduces a system architecture for the user intention realisation through the generation of code. The results employing the aforementioned architecture as well as a discussion is presented in \Cref{section:results}. Finally, the conclusion is presented in \Cref{section:conclusion}.

The implemented prototype as well as the generated codes and execution traces of different experiments can be found in our repository \footnote{\url{https://github.com/dos-group/LLMWorkflowGenerator/tree/hhai-2025}}.

%% file: section/related-work.tex
\section{Related Work}
\label{section:related-work}
The advent of LLMs and the public accessibility of ChatGPT \cite{wubriefgpt} have precipitated a surge in prospective applications, including those in the public health and medicine sector \cite{biswas2023role}\cite{thirunavukarasu2023large}\cite{nazi2024large}, the education and didactic sector \cite{firat2023chat}\cite{kasneci2023chatgpt}, and natural language processing in general \cite{kalla2023study}\cite{minaee2024large} with a growing necessity to categorize artificial intelligence \cite{morrisposition}. The transformer model \cite{vaswani2017attention} is the foundational concept behind LLMs \cite{chang2024survey}, which consequently, has resulted in the design and development of applications and frameworks such as AIOS \cite{mei2024aios}\cite{shi2024commands}, for machine-oriented user-intention resolution using LLMs. This domain originates from voice assistants, such as Siri, Cortana or Alexa \cite{hoy2018alexa}. Recent research around the idea of incorporating AI for resolving user intentions utilizes models which are trained for operating GUI applications \cite{liu2024autoglmautonomousfoundationagents}.

The generation of code with AI is a large subject of study. Evaluations show the feasibility of solving certain problems by generating code with GPT language models \cite{chen2021evaluating}\cite{lin2024soen101codegenerationemulating}. Tools like GitHub Copilot already support engineers in every day tasks \cite{MORADIDAKHEL2023111734}\cite{wermelinger2023}. Issues, such as hallucinations and erroneous code generation are addressed by fuzzing and static analysis \cite{Ouyang_2024}, novel approaches utilizing grammar augmentation \cite{ugare2024syncodellmgenerationgrammar} as well as the redesign of transformer decoding algorithms \cite{zhang2023planninglargelanguagemodels}.

%% file: section/preliminary-considerations.tex
\section{Preliminary Considerations}
\label{section:preliminary-considerations}
The process of rescheduling a meeting with a person, for example, involves a series of steps, including the retrieval of meeting information, the lookup of contact data, the initiation of communication with the person, the proposal of alternative dates, the negotiation with the person, and the consideration of proposed alternative dates by the person. This illustrates the complexity of generating such workflows. However, the effective generation of workflows is of paramount importance in the resolution of user intentions. In order to address intentions such as the aforementioned one, a semantic workflow model is required, which is capable of dealing with the complexity of such intentions.

From a macro perspective, a single workflow can be conceptualized as a finite state machine, constituted of multiple steps, which must be executed in a predetermined and ordered fashion for the resolution of a user intention. The interconnection of these steps allows for control flow with loops as well as case differentiations. Ideally, individual steps of these workflows are reusable throughout different workflows for reducing the overall complexity. The elements of workflows and their depiction as finite state machines, in conjunction with the employment of control flow with loops and case differentiations, directly result in the consideration and utilisation of programming languages in the context of workflows and their execution. LLMs can support this execution model by their ability to generate code.

%% file: section/method.tex
\section{Method}
\label{section:method}
\begin{figure}
    \centering
    \includegraphics[width=0.75\linewidth]{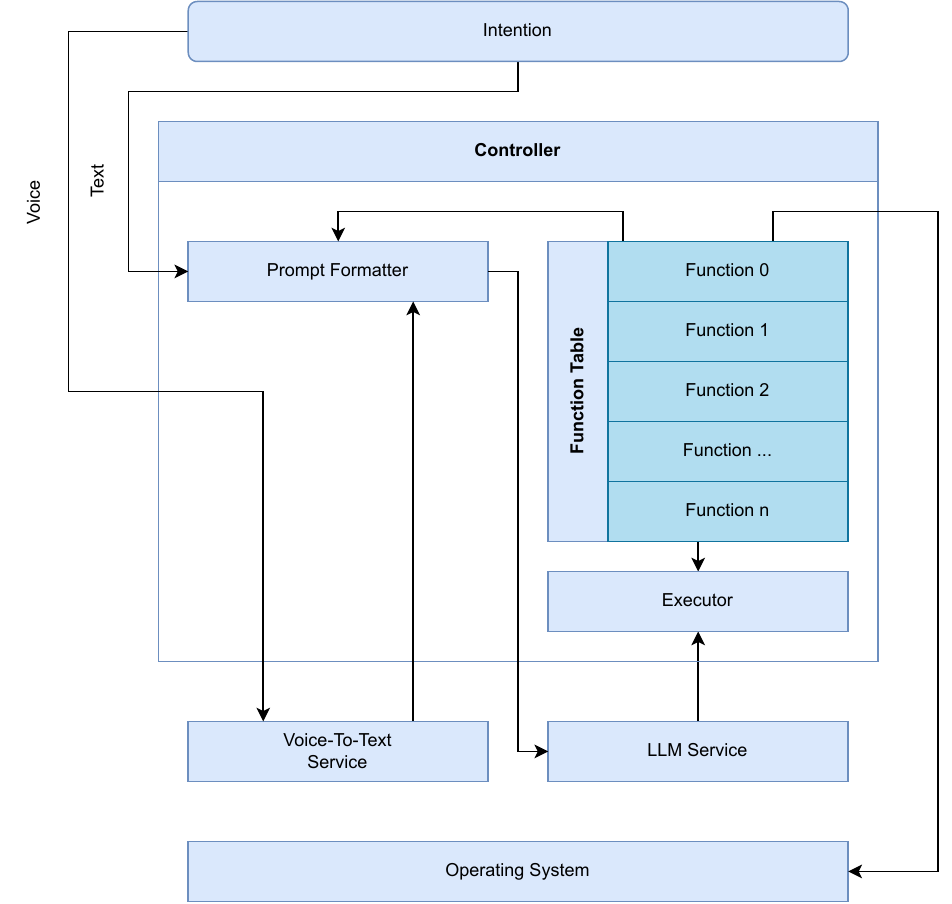}
    \caption{System architecture}
    \label{figure:system-architecture}
\end{figure}

We simulate the target environment of mobile devices by designing an architecture, as shown in \Cref{figure:system-architecture}, consisting of the following components:

\begin{itemize}
    \item \textbf{Operating system}
    \item \textbf{LLM Service} for generating code from a prompt
    \item \textbf{Voice-To-Text Service}
    \item \textbf{Controller} as intermediary component, including a \textbf{Prompt Formatter}, the \textbf{Function Table} and the \textbf{Executor}
\end{itemize}

In light of the inherent constraints of commodity hardware and the considerable resource demands of LLM inference, we propose the introduction of an \textbf{LLM Service} component, which is responsible for executing the LLM inference phase. The long-term objective of reducing the size of the LLM to a level that can be accommodated within the execution constraints of a typical mobile device represents a significant research challenge. In order to achieve this, it is essential to have a detailed understanding of the specific characteristics of the model in question. The LLM Service represents an abstraction for utilising the LLM, and it is designed to operate on an external infrastructure that is distinct from the user device.

The \textbf{Controller} serves as the central component responsible for addressing and encapsulating the LLM Service. It also oversees the serialisation and deserialisation of messages exchanged and manages the scheduling and execution of tasks. It is accessed by a text prompt, or by voice using a \textbf{Voice-To-Text Service} for transformation. 

Representing one of the most crucial components of this system, the \textbf{Function Table} exploits the ability of LLMs to generate imperative code consisting of functions as atomic, modular, and composable building blocks. This is analogous to a workflow comprising disparate steps. The Function Table serves two distinct purposes. Primarily, it is employed by the \textbf{Prompt Formatter} to generate a documentation of accessible functions, which is subsequently provided with the prompt to the LLM. Secondly, it contains the actual executable code of the specific function, which is utilised by the \textbf{Executor} for code execution following the generation step. These callbacks employ a range of operating system interfaces to facilitate a high-level application programming interface.

At the outset of the intention realisation process, the user inputs the intention. In the event that the intention is provided as voice input, the Controller is initially tasked with transforming the voice into text. Subsequently, the text is conveyed to the prompt formatter, which generates a prompt utilising a Prompt Template. This prompt format incorporates both the intention expressed in text and a documentation of the function signatures of the Function Table, thereby facilitating the LLM's comprehension of its capabilities. Subsequently, the prompt is conveyed to the LLM Service, which generates code based on the provided prompt. The Controller then reads the output and initiates the execution of the generated code.

We utilize the system architecture to demonstrate the feasibility of employing LLMs for the generation and execution of code based on user intentions for exploiting the generation of workflows. In these experiments, the user intentions are provided as text input to the Controller and are formatted with the prompt template illustrated in \Cref{figure:prompt-template}. The role of the LLM is set to \emph{You are a Python 3 code generator}. The Prompt Template is generated from the Function Table, which contains the signatures and implementation callbacks of the pertinent functions. The resulting output is then read and executed subsequently, utilizing Python's \emph{exec} function to execute the generated code with the designated functions in the global execution namespace. This configuration facilitates a constrained execution, whereby only the functions specified by the Function Table can be executed. In order to demonstrate the actual execution of the provided functions, the implementation callbacks utilize stub functions that print the name of the calling function and the arguments that have been passed. The following four intentions are considered:

\begin{enumerate}
    \item{\label{intention:1}\emph{Please send my car title to my insurance company}}
    \item{\label{intention:2}\emph{Please tell me the current temperature}}
    \item{\label{intention:3}\emph{Please play the song beat it by michael jackson}}
    \item{\label{intention:4}\emph{Please tell me all files in my home directory}}
\end{enumerate}

Each intention is fed to the system for a total of five times. An overview of the collected metrics, including the response time and the time to first token, is presented in \Cref{tabular:results}. For the first three intentions, the generated code successfully executes during the five trials. However, for intention \labelcref{intention:4}, two trials result in failure. The underlying causes of these two failures are attributed to two primary factors. Firstly, there was unauthorized access to packages that were not available in the isolated execution environment. Secondly, there was a violation of the rules governing scoping in programming languages. The intention with the most involved steps, intention \labelcref{intention:1}, is the intention for which the best resulting generated code is provided in Figure \Cref{figure:execution-trace} and its execution trace is shown in \Cref{figure:execution-trace}. The output for the remaining intentions can be located in the previously cited repository.

\begin{figure}
    \centering
    \input{figure/api_listing}
    \caption{Prompt template, which is generated from a function table.}
    \label{figure:prompt-template}
\end{figure}

\begin{table}[]
    \centering
    \begin{tabular}{c|c|c|c}
         Intention & Successes & Average Response Time (s) & Average Time to First Token (ms) \\
         \hline
         \labelcref{intention:1} & 5 & $\approx$ 3.31 & $\approx$ 466.0 \\
         \labelcref{intention:2} & 5 & $\approx$ 1.31 & $\approx$ 460.5 \\
         \labelcref{intention:3} & 5 & $\approx$ 2.03 & $\approx$ 491.4 \\
         \labelcref{intention:4} & 3 & $\approx$ 1.47 & $\approx$ 461.0 \\
    \end{tabular}
    \caption{Overview of the intentions and their average response time as well as their average time to first token for executing the particular generated code.}
    \label{tabular:results}
\end{table}

\begin{figure}
    \centering
    \input{figure/code_listing}
    \caption{Generated code sample for the intention \emph{Please send my car title to my insurance company}.}
    \label{figure:generated-code}
\end{figure}

\begin{figure}
    \centering
    \input{figure/trace_listing}
    \caption{Execution trace sample for the intention \emph{Please send my car title to my insurance company}.}
    \label{figure:execution-trace}
\end{figure}

%% file: figure/api_listing.tex
\begin{lstlisting}
You have the following application programming interface:

function find_file_id(expression: String): Integer|null
function find_contact_id(expression: String): Integer|null
function find_contact_email(contact_id: Integer):String|null
function play_voice(text: String): void
function ask_question(question: String): String
function play_audio_file(file: File): void
function send_email(email: String, subject: String, text: String, attachments: Collection<Integer>): void
function print_screen(text: String): void
function shell(command: String): String

Write Python 3 code only, which uses the application programming interface for the instruction
"{intention}"
\end{lstlisting}

%% file: figure/code_listing.tex
\begin{lstlisting}[language=Python]
# Step 1: Define the important variables
contact_email_expression = "insurance company"
email_subject = "Car Title Submission"
email_text = "Please find attached the car title for my vehicle."
attachments = []

# Step 2: Find the contact ID for the insurance company
contact_id = find_contact_id(contact_email_expression)

if contact_id is not None:
    # Step 3: Get the email address of the contact
    email_address = find_contact_email(contact_id)
    
    if email_address is not None:
        # Step 4: Find the car title file ID
        file_expression = "car title"
        file_id = find_file_id(file_expression)
        
        if file_id is not None:
            attachments.append(file_id)
        
        # Step 5: Send the email with car title attachment
        send_email(email_address, email_subject, email_text, attachments)
    else:
        print_screen("Could not find the email address for the insurance company.")
else:
    print_screen("Could not find the contact ID for the insurance company.")
\end{lstlisting}

%% file: figure/trace_listing.tex
\begin{lstlisting}
Execute "find_contact_id" and arguments "insurance company"
Execute "find_contact_email" and arguments "1"
Execute "find_file_id" and arguments "car title"
Execute "send_email" and arguments "john.doe@example.com", "Car Title Submission", "Please find attached the car title for my vehicle.", "[1]"
\end{lstlisting}

%% file: section/results.tex
\section{Results and Discussion}
\label{section:results}

The experiments presented illustrate the viability of using LLM technology for code generation and execution tasks, contingent upon the input of user intentions. GPT-4o-mini demonstrates remarkable proficiency in this domain. It is capable of comprehending both the previously unknown application programming interface and the specific user intention, subsequently generating accurate and well-behaved code that is mostly executable without any interference or issue. The generated code as well the execution traces reveal that the model is capable of comprehending not only the function signatures associated with the specified application programming interface, but also the intended purpose of each function solely based on its signature. Moreover, it is capable of accurately composing these functions, performing the requisite case differentiation, and providing a comprehensive explanation. For all provided examples, the LLM responds with the generated code within seconds. This remains within an acceptable range.

The primary objective for the generation of code is correctness. Our failed attempts in intention \labelcref{intention:4} underline this. The subsequent consideration is the duration of the generation. It raises the question of how specialized LLMs perform in this regard, which may involve further optimization through quantization, distillation, as well as pruning \cite{sreenivas2024llm}, and may ultimately enable the execution of LLMs on local devices, since in the experiments the LLM ran on external infrastructure and not locally. In general, building on well-established and well-understood operating system concepts with generic abstractions around multiuser and batch-oriented computing, as well as battle-tested technologies, appears to be a reasonable approach at first glance. Nevertheless, the domain of machine-oriented user intent realization is highly specialized and may not necessitate such comprehensive mechanisms. This prompts the question of how LLMs and their code generation capabilities might be employed in the future, as well as the extent to which the LLM-accessed interfaces will be either generic or specific.

The optimal use of the abilities of LLMs to generate code remains undetermined. Given its status as one of the most widely used programming languages in the current era and its extensive array of supporting tools and resources, as well as the fact that LLMs are primarily trained on publicly available data, we decided to use Python as our chosen programming language for code generation. Moreover, the provision of the \emph{exec} function facilitates the execution of code based on given functions within a restricted environment. Solutions based on serialisable message passing using exchange formats such as JSON were considered but ultimately rejected due to their lack of support for features such as control flow and function composition.

The automatic generation and execution of unverified code introduces a potential security hazard, which requires further investigation. From a technical vantage point, the isolated execution of unverified code necessitates counter-measures and hardening to prevent environment breakouts. It is imperative that the execution environment reside within a designated sandbox, thereby constraining its functionality to the parameters defined within the prompt. While soliciting user consent for the execution of generated code appears to be a viable solution, it ultimately necessitates the user's capacity for comprehending code or a graphical visualization of the generated code. However, this approach does not consider the potential for malicious uses such as code injections. A more viable approach would be to reduce the capabilities of functions that can be executed by the LLM. In the previously mentioned examples, this approach would entail the removal of the \textit{shell} function, as it provides a very powerful interface to the underlying operating system. Another viable mechanism involves the introduction of access layers that, upon access by the LLM, must be explicitly acknowledged by the user. Ensuring reproducibility and comprehensibility necessitates the implementation of a system capable of tracing the actions executed. The present system architecture is conducive to this objective by virtue of its provision of function wrappers.

The transition from manual operation of a GUI for user-intention resolution to the expression of user intention solely in natural language signifies a paradigm shift. Consequently, the transition toward automated resolution of user intentions gives rise to concerns regarding the reliance and dependability of users on AI and the impact of technology on users' understanding of the processes involved when resolving user-provided intentions manually. This issue is addressed in greater detail in \cite{Pfaltzgraf_Insch_2021}, which examines the technology gap that contemporary students face.

%% file: section/conclusion.tex
\section{Conclusions}
\label{section:conclusion}
This paper presents requisite actions for the resolution of user intentions. Moreover, the presented system architecture is employed to address workflows and their semantic aspects in experiments investigating the potential of machine-generated commands for the resolution of user intentions. The viability of employing LLM for the generation and execution of code is demonstrated. This is achieved by providing the LLM with prompts that include a user intention as well as a specification for an application programming interface. GPT-4o-mini displays a notable degree of precision in meeting this specific objective. The system is capable of comprehending the user intention and the proposed, yet unknown, application programming interface, effectively generating Python 3 code that is executed for the purpose of resolving exemplary user intentions. These findings suggest a promising future for the use of LLMs in this particular application domain. There is considerable potential for optimisation, particularly in the specific case of generating code from provided user intentions.

%% file: main.bbl
\begin{thebibliography}{10}

\bibitem{jin2024comprehensive}
Jin H, Zhang Y, Meng D, Wang J, Tan J.
\newblock A comprehensive survey on process-oriented automatic text summarization with exploration of llm-based methods.
\newblock arXiv preprint arXiv:240302901. 2024.

\bibitem{Carroll2017UnderstandingAR}
Carroll A.
\newblock Understanding and reducing smartphone energy consumption.
\newblock University of New South Wales; 2017.

\bibitem{achiam2023gpt}
Achiam J, Adler S, Agarwal S, Ahmad L, Akkaya I, Aleman FL, et~al.
\newblock Gpt-4 technical report.
\newblock arXiv preprint arXiv:230308774. 2023.

\bibitem{wubriefgpt}
Wu T, He S, Liu J, Sun S, Liu K, Han QL, et~al.
\newblock A Brief Overview of ChatGPT: The History, Status Quo and Potential Future Development.
\newblock IEEE/CAA Journal of Automatica Sinica. 2023;10(5):1122-36.

\bibitem{biswas2023role}
Biswas SS.
\newblock Role of chat gpt in public health.
\newblock Annals of biomedical engineering. 2023;51(5):868-9.

\bibitem{thirunavukarasu2023large}
Thirunavukarasu AJ, Ting DSJ, Elangovan K, Gutierrez L, Tan TF, Ting DSW.
\newblock Large language models in medicine.
\newblock Nature medicine. 2023;29(8):1930-40.

\bibitem{nazi2024large}
Nazi ZA, Peng W.
\newblock Large language models in healthcare and medical domain: A review.
\newblock In: Informatics. vol.~11. MDPI; 2024. p.~57.

\bibitem{firat2023chat}
Firat M.
\newblock How chat GPT can transform autodidactic experiences and open education? 2023.

\bibitem{kasneci2023chatgpt}
Kasneci E, Se{\ss}ler K, K{\"u}chemann S, Bannert M, Dementieva D, Fischer F, et~al.
\newblock ChatGPT for good? On opportunities and challenges of large language models for education.
\newblock Learning and individual differences. 2023;103:102274.

\bibitem{kalla2023study}
Kalla D, Smith N, Samaah F, Kuraku S.
\newblock Study and analysis of chat GPT and its impact on different fields of study.
\newblock International journal of innovative science and research technology. 2023;8(3).

\bibitem{minaee2024large}
Minaee S, Mikolov T, Nikzad N, Chenaghlu M, Socher R, Amatriain X, et~al.
\newblock Large language models: A survey.
\newblock arXiv preprint arXiv:240206196. 2024.

\bibitem{morrisposition}
Morris MR, Sohl-Dickstein J, Fiedel N, Warkentin T, Dafoe A, Faust A, et~al.
\newblock Position: Levels of AGI for Operationalizing Progress on the Path to AGI.
\newblock In: Forty-first International Conference on Machine Learning; 2024. Available from: \url{10.48550/arXiv.2311.02462}.

\bibitem{vaswani2017attention}
Vaswani A.
\newblock Attention is all you need.
\newblock Advances in Neural Information Processing Systems. 2017.

\bibitem{chang2024survey}
Chang Y, Wang X, Wang J, Wu Y, Yang L, Zhu K, et~al.
\newblock A survey on evaluation of large language models.
\newblock ACM Transactions on Intelligent Systems and Technology. 2024;15(3):1-45.

\bibitem{mei2024aios}
Mei K, Li Z, Xu S, Ye R, Ge Y, Zhang Y.
\newblock AIOS: LLM agent operating system.
\newblock arXiv e-prints, pp arXiv--2403. 2024.
\newblock Available from: \url{10.48550/arXiv.2403.16971}.

\bibitem{shi2024commands}
Shi Z, Mei K, Jin M, Su Y, Zuo C, Hua W, et~al.
\newblock From Commands to Prompts: LLM-based Semantic File System for AIOS.
\newblock arXiv preprint arXiv:241011843. 2024.

\bibitem{hoy2018alexa}
Hoy MB.
\newblock Alexa, Siri, Cortana, and more: an introduction to voice assistants.
\newblock Medical reference services quarterly. 2018;37(1):81-8.

\bibitem{liu2024autoglmautonomousfoundationagents}
Liu X, Qin B, Liang D, Dong G, Lai H, Zhang H, et~al.. AutoGLM: Autonomous Foundation Agents for GUIs; 2024.
\newblock Available from: \url{https://arxiv.org/abs/2411.00820}.

\bibitem{chen2021evaluating}
Chen M, Tworek J, Jun H, Yuan Q, Pinto HPDO, Kaplan J, et~al.
\newblock Evaluating large language models trained on code.
\newblock arXiv preprint arXiv:210703374. 2021.

\bibitem{lin2024soen101codegenerationemulating}
Lin F, Kim DJ, Tse-Husn, Chen. SOEN-101: Code Generation by Emulating Software Process Models Using Large Language Model Agents; 2024.
\newblock Available from: \url{https://arxiv.org/abs/2403.15852}.

\bibitem{MORADIDAKHEL2023111734}
{Moradi Dakhel} A, Majdinasab V, Nikanjam A, Khomh F, Desmarais MC, Jiang ZMJ.
\newblock GitHub Copilot AI pair programmer: Asset or Liability?
\newblock Journal of Systems and Software. 2023;203:111734.
\newblock Available from: \url{https://www.sciencedirect.com/science/article/pii/S0164121223001292}.

\bibitem{wermelinger2023}
Wermelinger M.
\newblock Using GitHub Copilot to Solve Simple Programming Problems.
\newblock In: Proceedings of the 54th ACM Technical Symposium on Computer Science Education V. 1. SIGCSE 2023. New York, NY, USA: Association for Computing Machinery; 2023. p. 172–178.
\newblock Available from: \url{https://doi.org/10.1145/3545945.3569830}.

\bibitem{Ouyang_2024}
Ouyang S, Zhang JM, Harman M, Wang M.
\newblock An Empirical Study of the Non-determinism of ChatGPT in Code Generation.
\newblock ACM Transactions on Software Engineering and Methodology. 2024 Sep.
\newblock Available from: \url{http://dx.doi.org/10.1145/3697010}.

\bibitem{ugare2024syncodellmgenerationgrammar}
Ugare S, Suresh T, Kang H, Misailovic S, Singh G. SynCode: LLM Generation with Grammar Augmentation; 2024.

\bibitem{zhang2023planninglargelanguagemodels}
Zhang S, Chen Z, Shen Y, Ding M, Tenenbaum JB, Gan C. Planning with Large Language Models for Code Generation; 2023.

\bibitem{sreenivas2024llm}
Sreenivas ST, Muralidharan S, Joshi R, Chochowski M, Patwary M, Shoeybi M, et~al.
\newblock LLM Pruning and Distillation in Practice: The Minitron Approach.
\newblock arXiv preprint arXiv:240811796. 2024.
\newblock Available from: \url{10.48550/arXiv.2408.11796}.

\bibitem{Pfaltzgraf_Insch_2021}
Pfaltzgraf D, Insch GS.
\newblock Digitally Native, Yet Technologically Illiterate: Methods to Prepare Business Students to Create Versus Consume.
\newblock Journal of Applied Business and Economics. 2021 May.
\newblock Available from: \url{https://articlegateway.com/index.php/JABE/article/view/4084}.

\end{thebibliography}
